\documentclass[conference]{IEEEtran}
\IEEEoverridecommandlockouts
\usepackage{cite}
\usepackage{amsmath,amssymb,amsfonts}
\usepackage{algorithmic}
\usepackage{graphicx}
\usepackage{textcomp}
\usepackage{xcolor}
\usepackage{flushend}
\usepackage{enumitem, url}
\usepackage{subcaption}
\usepackage{booktabs}
\usepackage{url}
\usepackage{multicol}
\usepackage{tikz}
\usetikzlibrary{shapes.geometric, arrows, positioning, fit}
\usepackage{multirow}
\def\BibTeX{{\rm B\kern-.05em{\sc i\kern-.025em b}\kern-.08em
    T\kern-.1667em\lower.7ex\hbox{E}\kern-.125emX}}
\usepackage[colorlinks,linkcolor=blue]{hyperref}

\title{Zero-TIG: Temporal Consistency-Aware Zero-Shot Illumination-Guided Low-light Video Enhancement\\
\thanks{This work was supported by the UKRI MyWorld Strength in Places Programme (SIPF00006/1) and EPSRC ECR (EP/Y002490/1).}}

\author{\IEEEauthorblockN{1\textsuperscript{st} Yini Li}
\IEEEauthorblockA{\textit{Visual Information Laboratory, University of Bristol} \\
Bristol, UK \\
ub24017@bristol.ac.uk}
\and
\IEEEauthorblockN{2\textsuperscript{nd} Nantheera Anantrasirichai}
\IEEEauthorblockA{\textit{Visual Information Laboratory, University of Bristol} \\
Bristol, UK \\
N.Anantrasirichai@bristol.ac.uk}
}

\begin{document}

\maketitle

\begin{abstract}
Low-light and underwater videos suffer from poor visibility, low contrast, and high noise, necessitating enhancements in visual quality. However, existing approaches typically rely on paired ground truth, which limits their practicality and often fails to maintain temporal consistency. To overcome these obstacles, this paper introduces a novel zero-shot learning approach named Zero-TIG, leveraging the Retinex theory and optical flow techniques. The proposed network consists of an enhancement module and a temporal feedback module. The enhancement module comprises three subnetworks: low-light image denoising, illumination estimation, and reflection denoising. The temporal enhancement module ensures temporal consistency by incorporating histogram equalization, optical flow computation, and image warping to align the enhanced previous frame with the current frame, thereby maintaining continuity. Additionally, we address color distortion in underwater data by adaptively balancing RGB channels. The experimental results demonstrate that our method achieves low-light video enhancement without the need for paired training data, making it a promising and applicable method for real-world scenario enhancement. Code is available at \href{https://github.com/liyinibristol/Zero-TIG}{https://github.com/liyinibristol/Zero-TIG}.

\end{abstract}

\begin{IEEEkeywords}
video enhancement, zero-shot learning, low-light, optical flow, underwater
\end{IEEEkeywords}

\section{Introduction}
\label{sec:intro}

Low-light conditions significantly challenge video quality. Camera sensors in dim environments capture fewer photons, crucial for image clarity and brightness, resulting in darker, less detailed footage. To compensate, cameras increase ISO sensitivity, which amplifies light signals but also heightens noise. Moreover, increasing exposure time can brighten videos but causes motion blur. These factors degrade color accuracy, contrast, and sharpness, making videos appear washed out and poorly defined. This issue is exacerbated in videos captured in deep water, where color distortion occurs due to wavelength-dependent light attenuation \cite{LU2023103926}. These distortions affect not only visibility, but also decision-making, and automation in many applications like security, surveillance, autonomous vehicles, medical imaging, and remote sensing. 

The performance of low-light \textit{image} enhancement has significantly improved with modern deep learning techniques, and research in this field continues to advance \cite{anantrasirichai:AI:2022}. However, applying these \textit{image}-based methods to low-light \textit{videos} introduces temporal flickering. Some supervised learning approaches extend to video by utilizing multiple frames as input \cite{Lin:ICIP:2024}, yet the effectiveness of low-light video enhancement remains limited. A major challenge is the lack of high-quality video-pair datasets, particularly for underwater scenes, which makes supervised deep learning limits its practicality. This underscores the need for self-supervised solutions to address these challenges effectively.

In this paper, we propose a novel self-supervised learning method, enabling high-quality video enhancement with only a single input video for training. Specifically, we introduce a novel zero-shot method, integrated with Retinex theory \cite{land1971lightness}, for real-world low-light video enhancement that simultaneously enhances contrast, reduces noise and corrects color at the same time. Additionally, our method recursively processes the reflectance and illumination components over time, ensuring temporal consistency and improved visual quality. In summary, the main contributions of this paper are as follows:
\begin{itemize}[noitemsep,topsep=0pt,leftmargin=10pt, itemindent=0pt]
\item We proposed a new zero-shot learning method, Zero-TIG, for low-light enhancement. It combines an enhancement module and a temporal feedback module, achieving visual quality without paired data.
\item We designed a feedback module that combining histogram equalization, optical flow (OF) estimation and image warping to recursively incorporate the enhanced results from the previous frame into the current pipeline, leading to less image noise and flickering.
\item For underwater data, we address the color distortion by computing the gain for each RGB channel. This ensures the illumination is constrained according to the average of individual channels and achieving the white balance automatically.

\end{itemize}

\begin{figure*}
    \centering
    \includegraphics[width=1\linewidth]{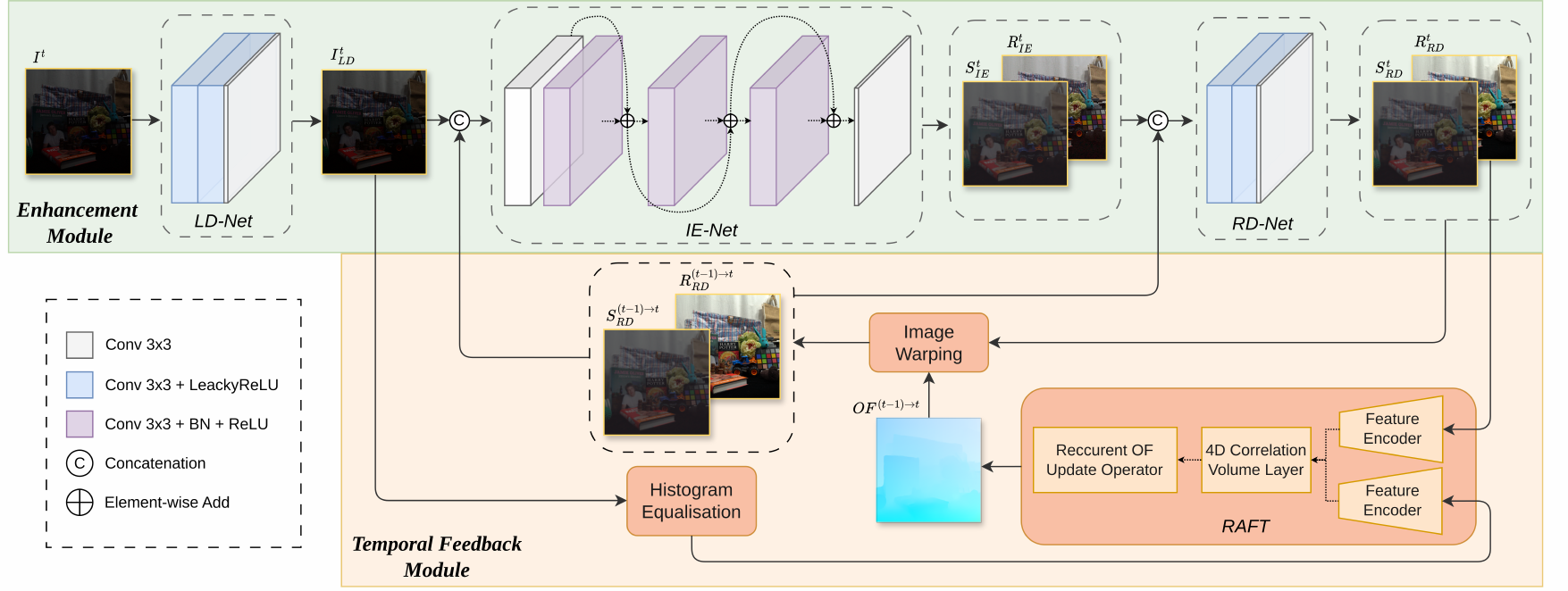}
    \caption{The proposed network includes an enhancement module with LD-Net, IE-Net, and RD-Net, and a temporal feedback module that employs histogram equalization, RAFT network for OF computation, and image warping. \(R^t_{RD}\) is the final output.}
    \label{fig:network}
\end{figure*}

\section{Related work}
\label{sec:related}

\subsection{Low-Light Video Enhancement}
Low-light video enhancement process typically begins with aligning feature maps of neighboring frames to the current frame \cite{wang2021sdsd}, followed by iterative feature re-weighting to reduce merging errors \cite{zhou2021rta}. Recent advances include a light-adjustable network based on Retinex theory \cite{Fu:dancing:2023} and a deep unfolding model using MAP optimization \cite{zhu2024unrolled}. State-of-the-art performance has been achieved by a wavelet conditional diffusion model \cite{Lin:Low:2024}, surpassing CNN- and transformer-based methods. Due to limited paired datasets, unpaired approaches such as CycleGAN \cite{anantrasirichai:Contextual:2021} and EnlightenGAN \cite{jiang2021enlightengan} have also been explored.

\subsection{Zero-Shot Learning for Enhancement}
Zero-shot learning addresses low-light enhancement without paired data. Zero-DCE \cite{Guo_2020_CVPR} introduced image-specific curve estimation using a lightweight network and non-reference losses. Subsequent work includes SGZSL \cite{Zheng:semantic:2022}, which integrates semantic segmentation, and RUAS \cite{liu2021ruas}, which combines Retinex theory with architecture search. \cite{Saeed:Lit:2023} proposed a three-stage model for iterative enhancement, while Zero-IG \cite{Shi:zero:2024} extended Noise2Noise \cite{Lehtinen:noise2noise:2018} for joint denoising and enhancement. Although \cite{Han:Exploring:2024} leverages temporal information to reduce flickering, its pixel-wise frame combination introduces ghosting artifacts. Notably, \cite{Zheng:semantic:2022} processes frames independently, limiting its applicability to video enhancement.

\section{Methodology}
\label{sec:method}

The proposed network is shown in Fig.~\ref{fig:network}, consisting of two main modules: the enhancement module and the novel temporal feedback module. The enhancement module is built on the basis of Zero-IG \cite{Shi:zero:2024}, and the temporal feedback module warps the previous enhanced output to the current frame, ensuring temporal consistency across the video sequence.

\subsection{Enhancement module}
Following \cite{Shi:zero:2024}, the enhancement module comprises three subnetworks which are the low-light denoising network (LD-Net), the illumination estimation network (IE-Net), and the reflection denoising network (RD-Net). 

To mitigate the impact of noise on the illumination distribution, the low-light image \(I^t\) at time \(t\) undergoes an initial denoising process. The LD-Net utilizes two downsamplers, \(G_1\) and \(G_2\), adopted from the framework proposed in \cite{Mansour:zero:2023} to generate a preliminary denoised image, denoted as \(I^t_{LP}\).
Inspired by Retinex theory, the IE-Net then decomposes \(I^t_{LP}\) into two components: reflectance \(R^t_{IE}\) and illumination \(S^t_{IE}\) as described by the following element-wise multiplication:
\begin{equation}
I^t_{LP}=R^t_{IE}\cdot S^t_{IE}\label{eq1},
\end{equation}
where \(R^t_{IE}\) and \(S^t_{IE}\) represent the intrinsic properties of the scene and the lighting conditions at time \(t\), respectively. To further enhance the performance of the denoising, the reflectance \(R^t_{IE}\) and illumination \(S^t_{IE}\) are concatenated and fed into the RD-Net. This subnetwork refines the output to produce \(R^t_{RD}\) and \(S^t_{RD}\), effectively removing residual noise and improving the overall quality of the enhanced image. The final output from the network is \(R^t_{RD}\), which represents the enhanced frame of the video.

\subsection{Temporal feedback module}
Although the enhancement network can achieve relatively good enhancement results, for videos, directly using single frames can lead to flickering. Consequently, we introduce the temporal feedback module to address the temporal consistency between consecutive frames in a video sequence.

The core concept of this module is to align the output result of the previous frame to the current frame, and then feed the aligned results back into IE-Net and RD-Net for continuous refinement. To achieve the alignment of the two frames, we used a highly efficient RAFT \cite{10.1007/978-3-030-58536-5_24} pre-trained model to calculate the optical flow (OF). Instead of directly applying the original noisy input \(I^t\), we utilize the pre-denoised image \(I^t_{LD}\), which reduces mismatches in optical flow computation caused by noise. Moreover, we apply histogram equalization to \(I^t_{LD}\) to align its intensity distribution with that of the enhanced frame, ensuring a more accurate optical flow estimation. We denote the histogram equalization process as \(HE(\cdot)\), and output as \(\hat{I}^t_{LD}\):
\begin{equation}
\hat{I}^t_{LD}=HE(I^t)\label{eq:HE}.
\end{equation}

Then, we input \(R^{t-1}_{RD}\), which is the final output of the network, and \(\hat{I}^t_{LD}\) into the RAFT network to compute the optical flow displacement map, denoted as \(OF^{(t-1) \rightarrow t}\):
\begin{equation}
OF^{(t-1) \rightarrow t}=RAFT( R^{t-1}_{RD},  \;  \hat{I}^t_{LD}) \label{eq:RAFT}
\end{equation}

In our experiments, we observed that weak texture regions in the images often lead to mismatches in the optical flow computation. To address this issue, we downsample both images by a factor of 3 along their height and width. This downsampling strategy not only mitigates the impact of noise, but also significantly accelerates the computation process. 

Based on \(OF^{(t-1) \rightarrow t}\), the image warping is implemented in both \(R^{t-1}_{RD}\) and \(S^{t-1}_{RD}\). These warped results, denoted as \(R^{(t-1) \rightarrow t}_{RD}\) and \(S^{(t-1) \rightarrow t}_{RD}\), are then fed into the IE-Net concatenated with \(I^t_{LD}\). Meanwhile, \(R^{(t-1) \rightarrow t}_{RD}\) and \(S^{(t-1) \rightarrow t}_{RD}\) are also fed into RD-Net concatenated with \(R^t_{IE}\) and \(S^t_{IE}\). For the first frame in a sequence, \(R^{(t-1) \rightarrow t}_{RD}\) and \(S^{(t-1) \rightarrow t}_{RD}\) are initialised as zero vectors. By using optical flow estimation and warping techniques, the context information from previous frames is fused to assist the denoising of the current frame and achieve smooth transitions.

\subsection{Loss functions}
Following \cite{Shi:zero:2024}, our method incorporates a total of 11 loss functions to optimize the model.
Inspired by \cite{Mansour:zero:2023}, the \(I^t_{LP}\) is downsampled into two subimages and is self-supervised by minimizing the difference between these subimages through a residual loss \(L_{res1}\) and a consistency loss \(L_{cons1}\). Given the two downsamplers denoted as \((G_1,G_2)\), the noise predicted by LD-Net as \(f_{LD}()\), and \(I\) as input \(I^t_{LD}\) in short, \(L_{res1}\) and \(L_{cons1}\) is described as:
\begin{align}
L_{res1} &= ||G_1(I) - f_{LD}(G_1(I)) - G_2(I)||^2_2 \nonumber \\
         &+ ||G_2(I) - f_{LD}(G_2(I)) - G_1(I)||^2_2\, \label{eq:L_res1}
\end{align}
\vspace{-0.5em}
\begin{align}
L_{cons1}&=||G_1(I) - f_{LD}(G_1(I)) - G_1(I - f_{LD}(I))||^2_2 \nonumber \\
          &+ ||G_2(I) - f_{LD}(G_2(I)) - G_2(I - f_{LD}(I))||^2_2\ , \label{eq:L_cons1}
\end{align}

Three constraints are applied for illumination estimation \(S^t_{IE}\). Specifically, the overall mean error loss \(L_{over}\) which computes the L2 loss between \(S^t_{IE}\) and a predefined brightness coefficient \(\alpha\) is leveraged:
\begin{equation}
L_{over}=||S^t_{IE} - \alpha^{-1} ||^2_2,\label{eq:L_over}
\end{equation}
where \(\alpha=0.5 Y^{-1}_L\) is set by \cite{Shi:zero:2024}. \(Y_L\) indicates the mean value of \(I^t_{LD}\).

To achieve amplitude adjustments for varying intensities, the pixel-wise adjustment loss \(L_{pix}\) is expressed as:
\begin{equation}
L_{pix}=||S^t_{IE} - \beta(\alpha I^t_{LD})^{\alpha} ||^2_2,\label{eq:L_pix}
\end{equation}
where the scaling factor is defined as \(\beta=\alpha^{-1}0.7^{-\alpha}\) in \cite{Shi:zero:2024}.

The illumination is additionally constrained by the smoothness loss \(L_{smooth}\), which regularizes the L1 loss of the horizontal and vertical gradient of \(S^t_{IE}\). In RD-Net, \(R^t_{RD}\) and \(S^t_{RD}\) are downsampled and optimized using a residual loss \(L_{res2}\) and a consistency loss \(L_{cons2}\). Additionally, \(S^t_{RD}\) is further constrained by an illumination consistency loss \(L_{ill}\) which is the mean squared error (MSE) between \(S^t_{RD}\) and \(S^t_{IE}\). The interactive denoising loss \(L_{inter}\) is used, which identifies noise by comparing the brightness channel differences between two downsampled versions of \(R^t_{RD}\). Furthermore, to refine the reflectance estimation, a local variance loss \(L_{var}\) and a color loss \(L_{color}\) are introduced to confine the variance and color change between \(R^t_{RD}\) and \(R^t_{IE}\).

\begin{figure}
    \centering
    \includegraphics[width=1\linewidth]{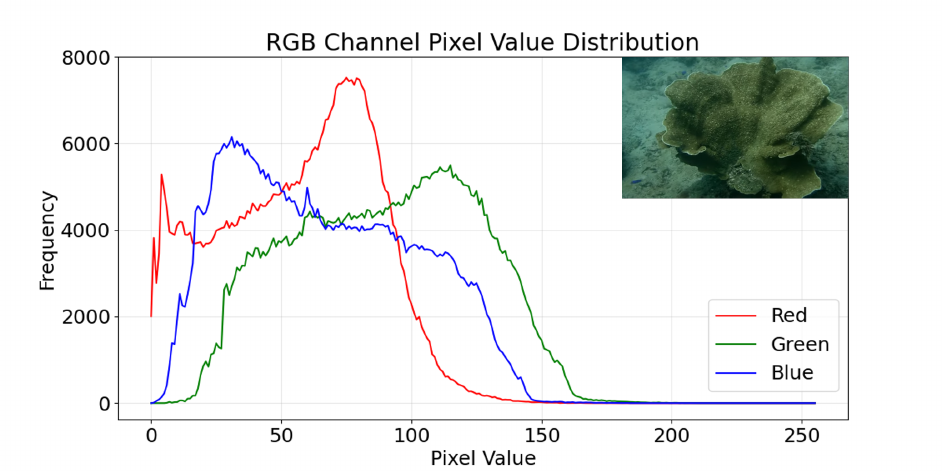}
    \caption{Frequency distribution of a underwater image.}
    \label{fig:rgbdistribution}
\end{figure}

\textbf{Adaptive white balance for underwater data.} The establishment of \(L_{over}\) and \(L_{pix}\) is based on the assumption that the distributions of the RGB channels of input are approximately uniform. However, as shown in Fig.~\ref{fig:rgbdistribution}, this assumption does not hold for underwater data because of the absorption and scattering of light in the water. Inspired by the Chromatic Retinex in \cite{Huang_Yang_Duan_Liu_2024}, we compute the mean value of luminance plane \(Y_L\) for each RGB channel independently, ensuring that all channels in \(S^t_{IE}\) are constrained individually. Accordingly, we extend (\ref{eq:L_over}) and (\ref{eq:L_pix}) as:
\begin{equation}
L_{over}=\sum_c||S^t_{IEc} - \alpha^{-1}_c ||^2_2,\label{eq:L_over_underwater}
\end{equation}
\vspace{-0.5em}
\begin{equation}
L_{pix}=\sum_c||S^t_{IEc} - \beta_c(\alpha_c I^t_{LD})^{\alpha_c} ||^2_2,\label{eq:L_pix_underwater}
\end{equation}
where \(c \in \{R,G,B\} \), and we empirically set \(\alpha_c=0.3Y^{-1}_{Lc}\), as the brightness of the underwater footage is generally higher than that of low-light test videos.

In conclusion, the final loss function is defined as \(L_{total}=L_{res1}+L_{cons1}+L_{over}+L_{pix}+L_{smooth}+L_{res2}+L_{cons2}+L_{ill}+L_{inter}+L_{var}+L_{color}\).

\begin{figure*}
    \centering
    \includegraphics[width=1\linewidth]{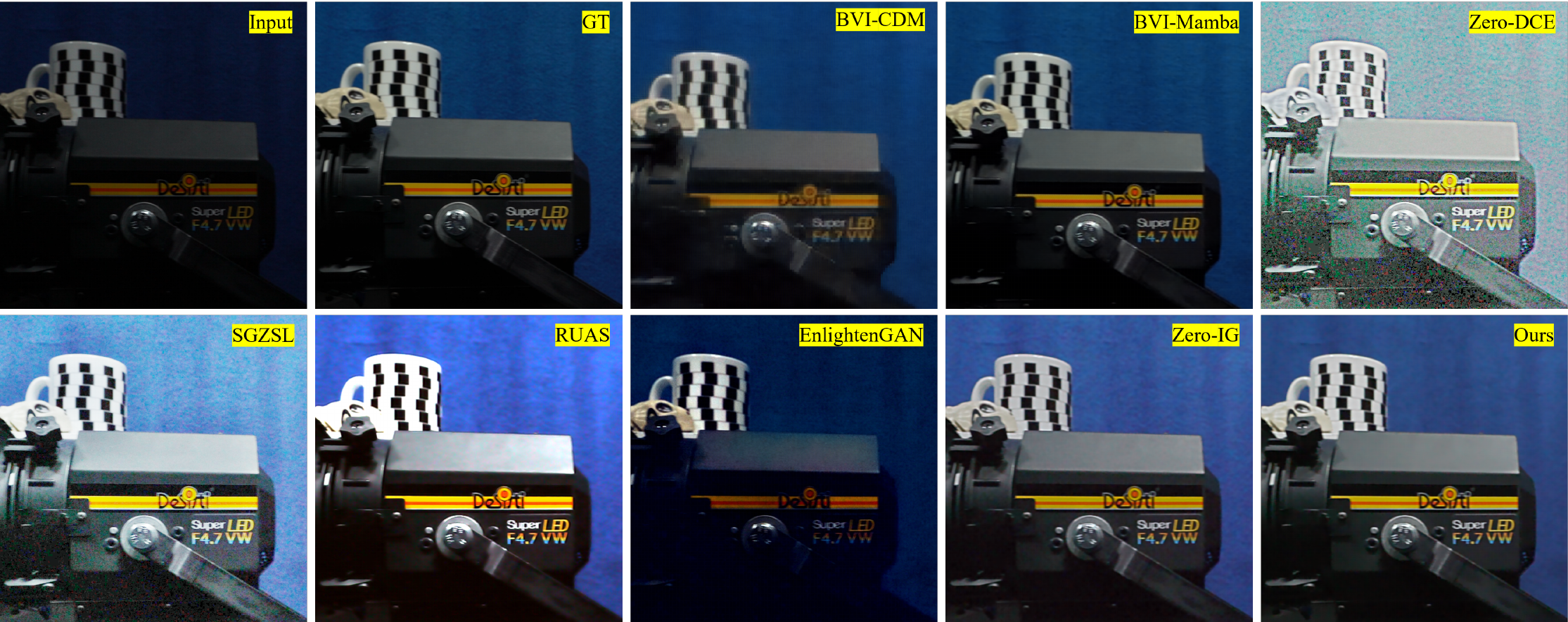}
    \caption{Comparison of visual results of our proposed Zero-TIG with supervised methods and self-supervised methods on BVI-RLV dataset. }
    \label{fig:cmp1}
\end{figure*}
\section{Experimental results and discussion}
\label{sec:results}

\subsection{Datasets and implementation}

The BVI-RLV dataset \cite{Lin:BVIRLV:2024} consists of paired videos featuring low-light footage and their corresponding normal-light versions. The sequences with normal light (100\%) are used as references for evaluation. We utilized 16 sequences from the BVI-RLV dataset to evaluate our proposed method. Each sequence is in RGB format (HD resolution, 25 fps). The contents are varied and dynamic, with different camera and object movement speeds.

We tested our proposed method on an underwater video from the BVI-Coral dataset \cite{Gough2025AquaNeRF}, which was captured at a depth of 10 meters, where the effects of low light and noise are present. 


The Adam optimizer was employed with \(\beta_1 = 0.9\), \(\beta_2 = 0.999\), a weight decay of \(3 \times 10^{-4}\) and a learning rate of \(10^{-4}\) on an NVIDIA V100 GPU. The RAFT network used a pre-trained model from the Sintel dataset \cite{Butler:ECCV:2012}, with frames downsampled by a factor of 3 to improve efficiency and accuracy. For BVI-RLV, Zero-TIG was initialized with a pre-trained weights on Zero-IG for 5 epochs and fine-tuned for another 5 epochs. For BVI-Coral, we modified the loss functions (\ref{eq:L_over}) (\ref{eq:L_pix}) to (\ref{eq:L_over_underwater}) (\ref{eq:L_pix_underwater}) and trained for 5 epochs.

\subsection{Performance for low-light video enhancement}

We compared the performance of our proposed Zero-TIG method with five state-of-the-art self-supervised learning methods for low-light enhancement: Zero-DCE \cite{Guo_2020_CVPR}, SGZSL \cite{Zheng:semantic:2022}, RUAS \cite{liu2021ruas}, Zero-IG \cite{Shi:zero:2024}, and EnlightenGAN \cite{jiang2021enlightengan}. Other methods proposed for video enhancement so far, e.g. \cite{zhu2024unrolled} and \cite{Han:Exploring:2024}, do not provide available codes, and we were unable to reproduce their results. We also include the results from supervised learning (BVI-CDM \cite{Lin:Low:2024} and BVI-Mamba \cite{huang2025bvi}) as upper-bound references. 

Table \ref{tab:comparison} presents the results, evaluated using well-known metrics: PSNR, SSIM, and LPIPS, as the BVI-RLV dataset provides normal light references. Also, since brightness can be subjective, we included results after applying histogram matching (HM) to the references to remove brightness differences, allowing us to evaluate the denoising performance of each method more accurately. These results demonstrate that our Zero-TIG outperforms most methods and is relatively comparable to Zero-IG. The EnlightenGAN is prone to generating hallucinated color artifacts in images, as subjectively evaluated and illustrated in Fig. \ref{fig:cmp1}. Compared to Zero-IG, our method demonstrates superior denoising performance. Although supervised methods achieve higher quantitative metrics, our visual results are comparable to, and in some cases even surpass, those of supervised approaches, outperforming BVI-CDM in terms of visual quality.

\begin{figure} 
    \centering
    \includegraphics[width=0.85\linewidth]{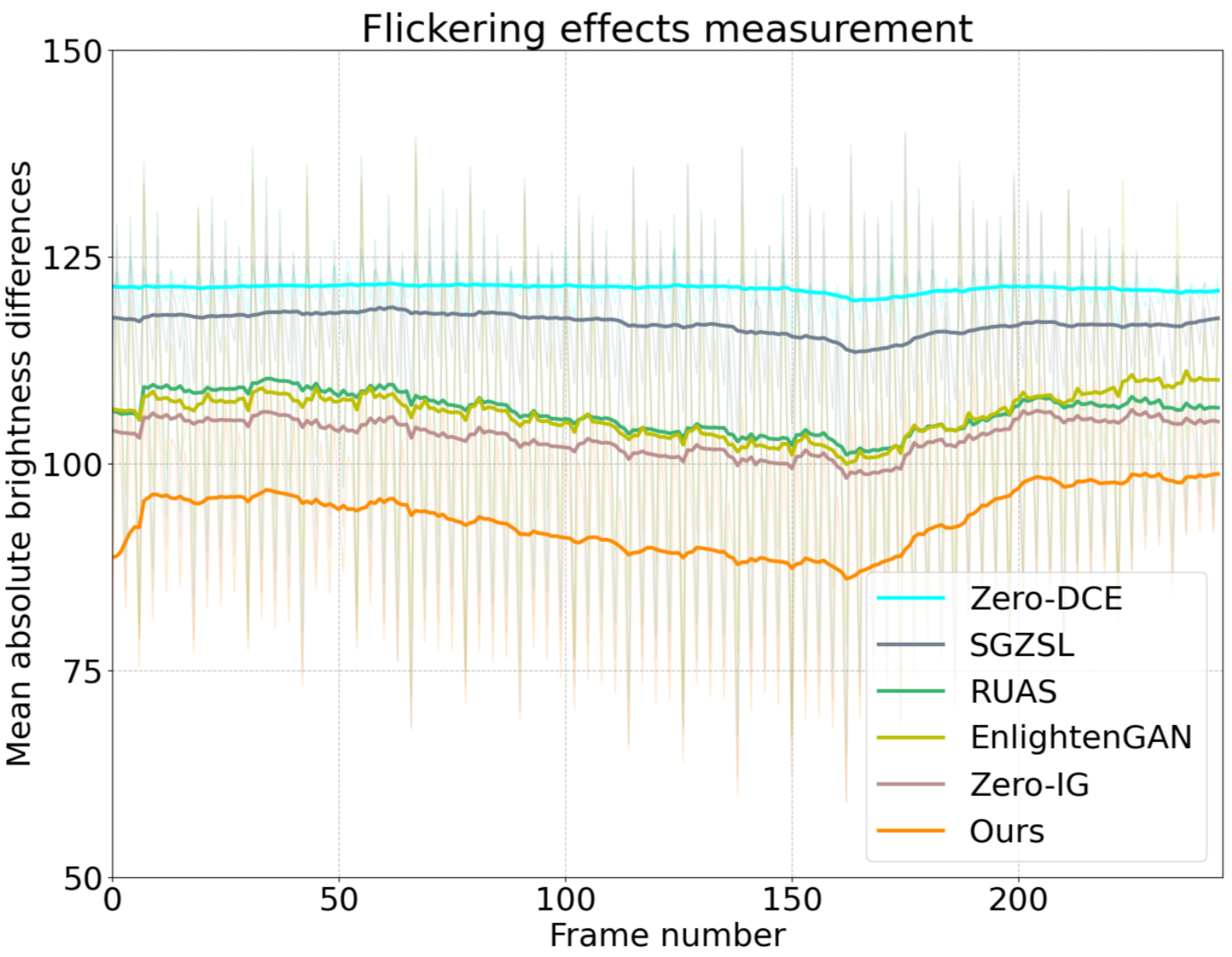}
    \caption{MABD vectors of Zero-IG and our method. Moving average method is applied to data for clarity.}
    \label{fig:mabd}
\end{figure}

We also evaluated temporal consistency using Mean Absolute Brightness Differences (MABD), as proposed in \cite{Jiang:learn:2019}. A lower MABD value indicates better sequential continuity. For comparative analysis, we selected a representative video named \textit{S11\_gift\_wrap} with 10\% illuminance and computed the MABD values across consecutive frames. To enhance clarity, we applied a moving average technique with a window size of 15. Fig. \ref{fig:mabd} illustrates the flickering effects of different methods after enhancement. Our approach exhibits markedly smaller amplitude, indicating more stable brightness variations over time. which confirm that our method demonstrates the superior temporal coherence of our method compared to Others. 




\begin{table}[t]
    \centering
    \caption{Comparison of PSNR, SSIM, and LPIPS across different methods with and without histogram matching (HM)}
    \resizebox{\columnwidth}{!}{
    \begin{tabular}{lcccccc}
        \hline
        & \multicolumn{2}{c}{PSNR} & \multicolumn{2}{c}{SSIM} & \multicolumn{2}{c}{LPIPS} \\
        & w/o HM & w/ HM & w/o HM & w/ HM & w/o HM & w/ HM \\
        \hline
        \textcolor{gray}{BVI-CDM}$^*$ & \textcolor{gray}{30.51} & \textcolor{gray}{-} & \textcolor{gray}{0.888} & \textcolor{gray}{-} & \textcolor{gray}{0.089} & \textcolor{gray}{-}\\
        \textcolor{gray}{BVI-Mamba}$^*$ & \textcolor{gray}{31.22} & \textcolor{gray}{-} & \textcolor{gray}{0.912} & \textcolor{gray}{-} & \textcolor{gray}{0.071} & \textcolor{gray}{-} \\
        \hline
        Zero-DCE & 10.540 & 18.932 & 0.430 & 0.488 & 0.528 & 0.507 \\
        SGZSL & 13.416 & 24.026 & 0.577 & 0.723 & 0.420 & 0.380 \\
        RUAS & 15.305 & 18.520 & 0.631 & 0.712 & 0.481 & 0.515 \\
        EnlightenGAN & 15.486 & 17.875 & 0.518 & 0.550 & 0.515 & 0.522 \\ 
        Zero-IG & \textbf{19.374} & \underline{27.840} & \underline{0.639} & \underline{0.834} & \underline{0.398} & \underline{0.370} \\
        \hline
        Zero-TIG (ours) & \underline{19.340} & \textbf{28.052} & \textbf{0.790} & \textbf{0.854} & \textbf{0.360} & \textbf{0.368} \\
        \hline
        \multicolumn{5}{l}{$^*$ Supervised learning methods} \\
    \end{tabular}}
    \label{tab:comparison}
\end{table}

\subsection{Performance of low-light underwater video enhancement}

\begin{figure} 
    \centering
    \includegraphics[width=1\linewidth]{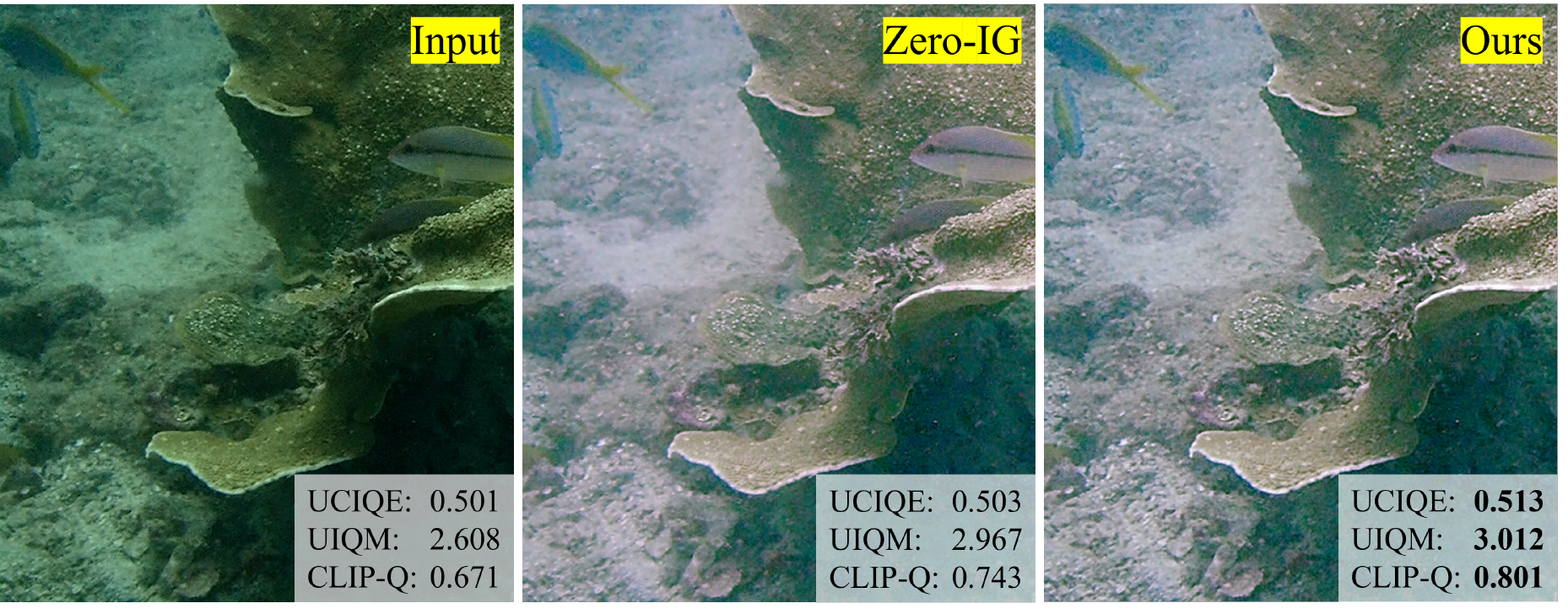}
    \caption{Results of underwater video enhancement}
    \label{fig:underwaterresults}
\end{figure}

In this experiment, we compared our method with Zero-IG. Since ground truth for the underwater scene is unavailable, we used non-reference metrics and subjective assessment. The evaluation metrics include two underwater evaluation metrics UIQM \cite{7305804} and UCIQE \cite{7300447}, and the quality score in CLIP-IQA \cite{wang2022exploring}. Higher values of these metrics indicate better image quality. We selected a representative video, \textit{Original210120.MOV}, from \cite{Gough2025AquaNeRF} for evaluation and visualized the results of one frame, as shown in Fig. \ref{fig:underwaterresults}.

Our results demonstrate superior contrast compared to Zero-IG, with clearer edges of coral and sand in the background, as illustrated in Fig. \ref{fig:underwaterresults}. Quantitatively, our method achieves higher scores across all non-reference metrics. These improvements confirm that our approach effectively addresses distortions in underwater videos and achieves adaptive white balance, enhancing both visual quality and metric performance.

\section{Conclusion}
\label{sec:conclusion}

This paper proposes Zero-TIG, a zero-shot self-supervised method for low-light video enhancement that integrates Retinex theory with a novel temporal feedback module to reduce flickering and noise. Additionally, an adaptive white balance is introduced for underwater data by constraining RGB channels of illumination, achieving superior temporal coherence and color accuracy without paired training data.

\bibliographystyle{IEEEtran}
\bibliography{bib}

\end{document}